\documentclass[10pt, a4paper]{article}
\usepackage{lrec}
\usepackage{multibib}
\newcites{languageresource}{Language Resources}
\usepackage{graphicx}
\usepackage{tabularx}
\usepackage{soul}
\usepackage{url}

\usepackage{epstopdf}

\usepackage{hyperref}
\usepackage{xstring}

\usepackage{amsmath,bm}
\usepackage{booktabs}
\usepackage{multirow}
\usepackage{float}

\usepackage{CJKutf8}

\title{JESC: Japanese-English Subtitle Corpus}

\name{Reid Pryzant$^1$, Youngjoo Chung$^2$, Dan Jurafsky$^1$, Denny Britz$^3$}

\address{$^1$Stanford University, $^2$Rakuten Institute of Technology, $^3$Google Brain \\
         \{rpryzant, jurafsky\}@stanford.edu, yjchung@acm.org, dennybritz@gmail.com}

\abstract{
In this paper we describe the Japanese-English Subtitle Corpus (JESC). JESC is a large Japanese-English parallel corpus covering the underrepresented domain of conversational dialogue. It consists of more than 3.2 million examples, making it the largest freely available dataset of its kind. The corpus was assembled by crawling and aligning subtitles found on the web. The assembly process incorporates a number of novel preprocessing elements to ensure high monolingual fluency and accurate bilingual alignments. We summarize its contents and evaluate its quality using human experts and baseline machine translation (MT) systems. 
 \\ \newline \Keywords{parallel corpus, asian languages, machine translation} }

\begin{document}

\maketitleabstract

\section{Introduction}

There is a strong need for large parallel corpora from new domains. Modern machine translation (MT) systems are fundamentally constrained by the availability and quantity of parallel corpora. Apart from the exceptions of English-Arabic, English-Chinese, and several European pairs, parallel corpora remain a scarce resource due to the high cost of manual construction \cite{chu2014constructing}. Furthermore, despite promising work in domain adaptation, MT systems struggle to generalize to new domains that are disparate from their training data \cite{pryzanteffective}.

This need for large, novel-domain data is especially evident in the resource-poor Japanese-English (JA-EN) language pair. Only two large ($>$1M phrase pairs) and free datasets exist for this language pair \cite{phontron,opus,moses}. The first is called ASPEC. It consists of 3M examples and it originates from scientific papers, a highly formalized and written domain (all other JA-EN datasets have similar language) \cite{nakazawa2016aspec}. The other, OpenSubtitles, is a multi-language dataset of aligned subtitles authored by professional translators; the JA-EN subset of these data contains approximately 1M examples \cite{lison2016opensubtitles2016}. OpenSubtitles is to the best of these authors knowledge the \emph{only} parallel corpus to cover the unrepresented domains of conversational speech and informal writing. This dearth of large-scale and informal data is especially problematic because colloquial Japanese has significant structural characteristics which can preclude cross-domain translation \cite{tsujimura2013introduction}. We hope to alleviate this problem by building off the work of  \cite{lison2016opensubtitles2016} to construct a larger corpus that incorporates the vast number of unofficial and fan-made subtitles on the web. 

Subtitles are an excellent source for alleviating resource scarcity problems. There are a wide and interesting range of linguistic phenomena in subtitles that are poorly represented elsewhere. This includes colloquial exchange, slang, expository discourse, dialect, vernacular, and movie dialog, which is available in great quantities and
has been shown to resemble natural conversation \cite{fochini13}. Furthermore, large subtitle databases are freely available on the web, are often crowd-sourced, and the close correspondence between subtitles and their video material renders time-based alignment feasible \cite{tiedemann2008synchronizing}. 

We release JESC, a new Japanese-English parallel corpus consisting of 3.2 million pairs of crawled TV and movie subtitles\footnote{The dataset and code are available at \\\url{https://nlp.stanford.edu/projects/jesc/}}. We also release the tools, crawlers, and parsers used to create it.  We provide a comprehensive statistical summary of their contents as well as strong baseline machine translation systems that yield competitive BLEU scores. This is the largest freely available Japanese-English dataset to date and covers the resource-poor domain of conversational or informal speech.

\section{Source Data}

We crawled four free and open subtitle repositories for Japanese and English subtitles: \texttt{kitsunekko.net}, \texttt{d-addicts.com}, \texttt{opensubtitles.org}, and \texttt{subscene.com}. Each subtitle database accepts submissions from the public and disseminates them through a web interface. There is no standard imposed on subtitle submissions, and as such, they exist in a plenitude of file formats, encodings, languages (beyond that being advertised), and content (beyond that being advertised). Though some of these subtitles are indeed the ``official'' translation, many were translated or transcribed by amateur fans of the video content. Thus, many of our translations are crowd-sourced, and there are no guarantees on the fluency of the participants. Many subtitle files contained grammatical, spelling, optical character recognition (OCR), and a host of other problems that preclude their direct usage for machine translation. 

\begin{figure}[H]
	\centering
	\includegraphics[width=0.5\textwidth]{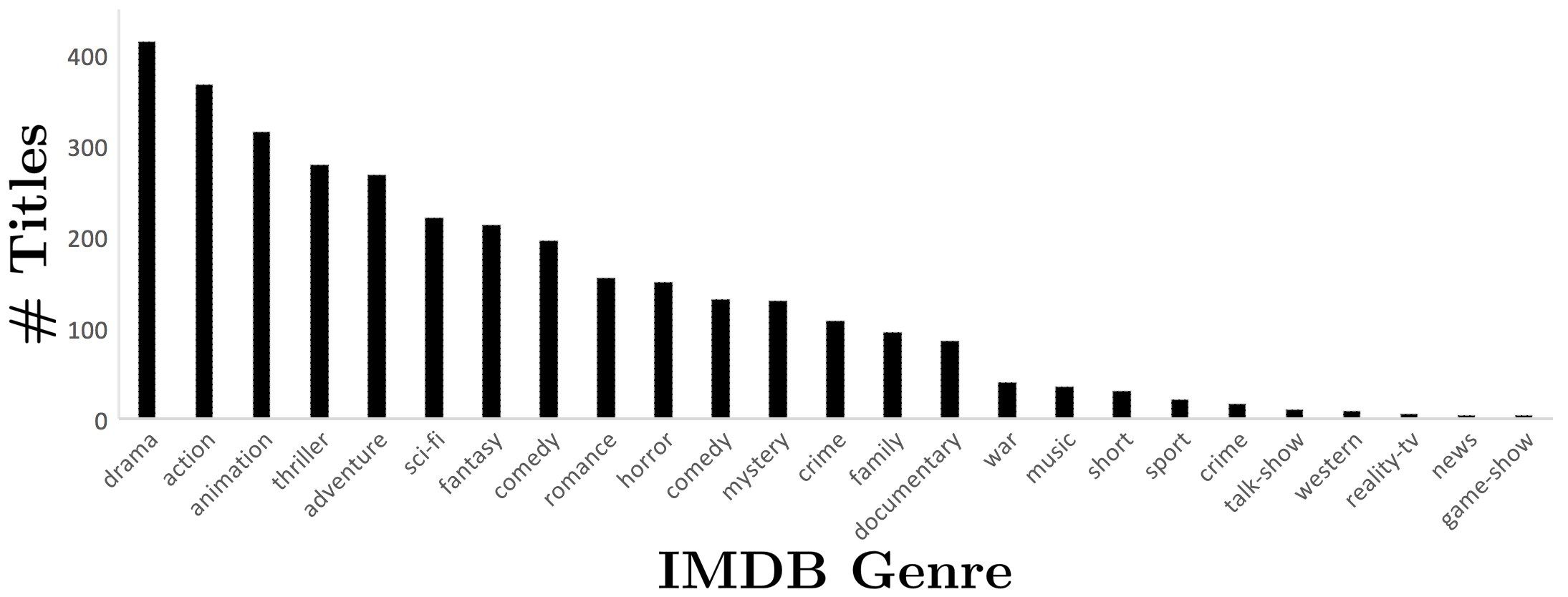}
	\caption{Genre distribution for our crawled titles (obtained via IMDB).}
    \label{fig:genres}
\end{figure}

Crawling these online repositories yielded 93,992 subtitle files corresponding to 23,318 individual titles (episodes, etc.), 4,610 grouped titles (shows, etc.), and more than 100 million individual captions corresponding to a broad range of video material (Figure \ref{fig:genres}). Our objective is to automatically cull a high quality parallel corpus from this unstructured and error-prone data.

\section{Preprocessing}

\begin{figure*}[t]
\includegraphics[width=1\linewidth]{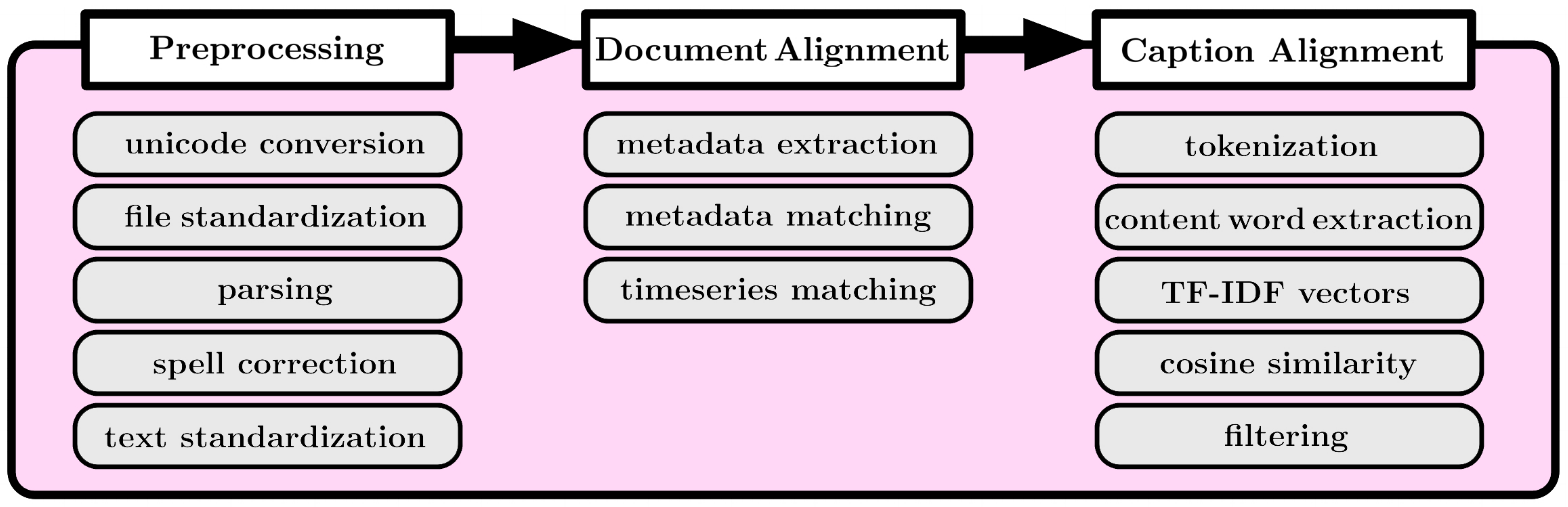}
\caption{Workflow for the creation of parallel corpora from raw subtitle files.}
\label{figure:model}
\end{figure*}
\label{sec:mining}

Due to the acute heterogeneity and high error rate of our subtitle files, we underwent a number of preprocessing steps to bring them into a form suitable for alignment. The output of this preprocessing pipeline is a series of documents (one per subtitle file), each structured as a titled list of captions, start times, and end times.

\subsection{Document Standardization}

First, we converted each subtitle file into a standardized format. We applied the \texttt{chardet} library to determine the most likely encoding \cite{li2001composite}, and converted this encoding to a \texttt{utf-8} standard. We then used the \texttt{ffmpeg} library to convert files into a common SubRip (\texttt{.srt}) format \cite{tomar2006converting}. Files that \texttt{ffmpeg} was unable to convert were interpreted as illegitimate and discarded. Last, we parsed these \texttt{.srt} documents into machine-readable YAML\footnote{\url{http://www.yaml.org/}}. Each resulting document contains a title (obtained from \texttt{.srt} metadata) and a list of captions, with each caption consisting of tokenized body text, start time, and end time.  

\subsection{Text Correction}

Next, we preprocessed the English documents by performing syntax correction on each caption. Many fan-made subtitles were created by non-native English speakers and as such contained typographical and spelling mistakes. We developed a laplace-smoothed statistical error model $P(w \vert w^*)$ that scores the probability of a word $w^*$ being misspelled as $w$. This model was trained by observing relative misspelling frequencies on the Birkbeck corpus \cite{mitton1985birkbeck}. We then developed two additional laplace-smoothed frequency-based models using unigrams and bigrams from Google's Web 1T N-grams \cite{islam2009real}. These are language models that score the prior probability of n-gram occurrence, $P(w)$, and the transition probability $P(w_i \vert w_{i-1})$. We used a smoothing factor of $\alpha = 1$ for all of these models. Next, for each possibly misspelled token $t_i$ of a caption $c$, we performed depth-4 uniform cost search on the space of edits to produce candidate replacements $t^*_i$. Armed with the error model $P(t_i \vert t^*_i)$ and language model $P(t^*_i) P(t^*_i \vert t_{i - 1})$, we scored the probability of each candidate by applying Bayes rule, similar to \cite{lison2016opensubtitles2016}:

\[P(t^*_i \vert t_i, t_{i - 1}) = P(t_i \vert t^*_i) P(t^*_i) P(t^*_i \vert t_{i - 1}) \]

Note that this checker improves on that of \cite{lison2016opensubtitles2016} with the inclusion of a data-driven error model, prior term, and depth-4 uniform cost search (as opposed to making any correction with $>$50\% probability).

We also standardized the text of each caption by lowercasing, removing bracketed text, out-of-language subsequences (e.g. encoding errors, OCR errors, machine-readable tags), linguistic queues (i.e. ``laughs''), inappropriate punctuation (e.g. leading dashes, trailing commas), and author signatures.

\section{Cross-lingual Alignment}

Once these subtitle files are brought into a suitable form, they can be aligned to form a parallel corpus. Doing so requires alignment at two levels: the document level, where we group subtitles according to the movie or TV show they correspond too, and the caption level, where we determine which captions are direct translations of one another.

\subsection{Document Alignment}

In order to align subtitles across distinct languages we must first align the documents themselves, i.e. determine which subtitle documents' captions are worth aligning. This is because (1) multiple subtitle documents may exist for a given movie or TV episode, and (2) subtitles from non-matching movies or TV shows will not be in correspondence.

We generated candidate alignments between Japanese and English documents with a novel technique involving soft matching on file metadata. We first extracted metadata in the form of movie and TV show names as well as episode numbers from each document title. Next, we used the Ratcliff-Obershelp algorithm to determine pairwise title similarities (this algorithm determines similarity via the lengths of matching subsequences), matching two files if their similarity ratio exceeded 90\% \cite{ratcliff1998ratcliff}. We proceeded to filter out pairs with differing episode numbers. 

We refined document alignments with another novel method which considers the temporal sequence of their captions. We created document vectors $D_i = [d_1^i, ..., d_{10,000}^i]$ for each subtitle file $i$. Each feature $d_k^i$ is a binary indicator that is active when document $i$ has a caption whose closest starting second is $k$. To account for possible time shift errors, we constructed a multiplicity of vectors for each document, each shifted to a different start time. We then computed the Hamming distance between each Japanese-English document vector and discarded those pairs with a distance greator than 0.04 (chosen based on a bucketed distribution of distances between all pairs).

\subsection{Caption Alignment}
Now that we have a set of matched English and Japanese subtitle files $\{(\bm{E}_1, \bm{J}_1), ..., (\bm{E}_n, \bm{J}_n)\}$, we must align the captions of each pair such that captions which are direct translations of each other are selected for extraction.

Let $\bm{E} = e_1, ..., e_n$ and $\bm{J} = j_1, ..., j_m$ be a pair of aligned English and Japanese documents that presumably map to similar video content. Note that each $e_i$ and $j_i$ are subtitle captions consisting of a start time $a_i$, end time $b_i$, and a sequence of text tokens $t_1, ..., t_z$ (in Japanese or English). If $\bm{E}$ and $\bm{J}$ were in perfect harmony then we would be able to pair $e_1$ with $j_1$, $e_2$ with $j_2$ and so on. However, matched documents are rarely in such close correspondence. Optical Character Recognition (OCR) errors, misaligned files, differing start times, speed ratios, framerate, and a host of other problems preclude such a one-to-one correspondence \cite{tiedemann2016finding}. 

Due to the severity of the aforementioned problems, especially among documents that have been subtitled by amateur translators, we found existing caption alignment algorithms inadequate for our needs. We developed a novel subtitle alignment algorithm that matches captions based on both timing and content. For each Japanese caption, we search a nearby window (typically 10-15 seconds) of English captions and score their similarity. We then take the highest-scoring match of this window. 

We score the quality of an English-Japanese caption pairing by (1) morphologically analyzing Japanese and English captions and discarding all but the content words, then (2) stemming these content words, (3) translating the Japanese to English with simple dictionary lookups, (4) averaging the GLoVE vectors for each caption's words, and (5) computing the cosine similarity between these vector representations. We used the Rakuten and JUMAN morphological analyzers to extract content words from Japanese captions, and the Stanford POS tagger for English \cite{hagiwara2014lightweight,manning2014stanford}. We used JUMANPP \cite{morita2015morphological} and NLTK to stem these words \cite{bird2006nltk}, and JMdict/EDICT to map Japanese words to their English equivalents \cite{breen2004edict,matsumoto1991user}. Phrases without translations were skipped. Note that our method introduces a bias in the phrase pairs of resultant matches, namely those pairs that would score highly under a lexicon, but we assume that JMdict/EDICT is near-complete with respect to common content words.

\subsection{Filtering}

The document- and caption-matching procedures outlined above produced 27,716,868 matches between English and Japanese captions. We proceeded by filtering out low-quality matches, choosing to retain only the very highest quality matches. We discarded matches whose cosine similarity was below the $84^{\mathrm{th}}$ percentile (assuming a normal distribution), leaving 4,434,699 pairs. This percentile was chosen based on downstream NMT performance. Last, we removed duplicate matches and out-of-language matches (matches where $< 90\%$ of the characters in $e$ or $>10\%$ of the letters in $j$ are roman), leaving us with a final count of 3,240,661 phrase pairs.

\section{Investigation}

\begin{CJK}{UTF8}{min}
\begin{table*}[t]
\centering
\label{my-label}
\begin{tabular}{@{}ll@{}}
\toprule
English & Japanese \\ \midrule
look, i don’t do that shit anymore. & 私は卒業した \\ \\
thank you! & ありがとう \\ 
you're so sweet &  \\ \\
look, his name is cyrus gold. & いいか 彼の名前はサイラス・ゴールド \\ \\
is that so? i hate to disappoint you. & そうか それは残念だったな。 \\ \bottomrule
\end{tabular}
\caption{Samples from JESC.}
\end{table*}
\end{CJK}

\subsection{Basic statistics}
\begin{figure}[]
	\centering
	\includegraphics[width=0.5\textwidth]{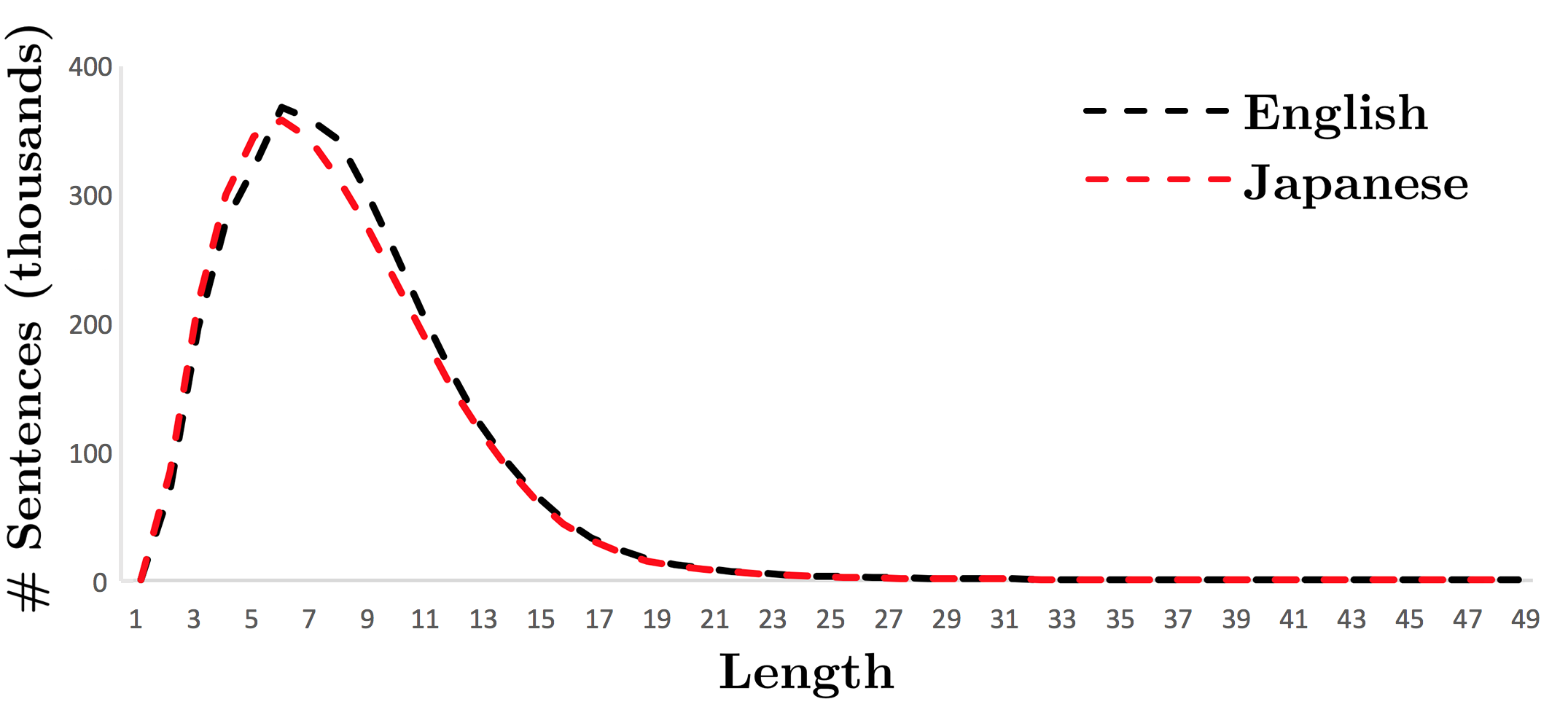}
	\caption{JESC exhibits a right-skewed sentence length distribution. 83 English and 114 Japanese phrases have length $>50$ .}
	\label{fig:sent-len-dist}
\end{figure}

The resulting corpus, which we call JESC, for Japanese English Subtitle Corpus, consists of 29,368 unique English words and 87,833 unique Japanese words. The train/val/test splits are of size 3,236,660\ /\ 2000\ /\ 2001. The lengths of each languages' phases are quite similar (Figure \ref{fig:sent-len-dist}). JESC consists mainly of short bursts of conversational dialogue; the average English sentence length is 8.32 words while for Japanese it is 8.11. 

JESC also exhibits multiple reference translations for 163,665 and 130,790 Japanese and English phrases, respectively. For example, the English sentence ``what?'' has translations such as \begin{CJK}{UTF8}{min}何だ?/なんだって？/な　なんだよ？/どうしたんですか?\end{CJK} due to variations in the Japanese suffix determined by the circumstances of the speaker and dialogue situation. This feature makes it unique among large parallel corpora and greatly improves its usefulness. While BLEU is designed to benefit from multiple reference translations \cite{Papineni02bleu}, this is a luxury rarely afforded to the modern system, and both of the major MT workshops use single-reference BLEU to evaluate all their tasks\footnote{\url{http://www.statmt.org/wmt17/}}\footnote{\url{http://lotus.kuee.kyoto-u.ac.jp/WAT/WAT2017/}}. 



\subsection{Evaluation}
\subsubsection{Alignment evaluation}
We checked the validity of bilingual sentence alignments based on  the procedure of \cite{utiyama2007japanese}. A pair of human evaluators (both native Japanese and proficient English speakers) randomly sampled 1000 phrase pairs. On average, 75\% of these pairs were perfectly aligned, 13\% partially aligned, and 12\% misaligned. There was strong agreement between these adjudicators' findings (Cohen's kappa of 0.76) so we may conclude that JESC is noisy but has significant signal that can be useful for downstream applications. 


\subsubsection{Translation evaluation}

In addition to alignment, we evaluated the quality of crowd-sourced translation. Our evaluators used the Japanese Patent Office's adequacy criterion (JPO). The JPO is a 5-point system which provides strong guidelines for scoring the quality of a Japanese-English translation pair \cite{nakazawa2016aspec}. Again in the style of \cite{utiyama2007japanese} we sampled and evaluated 200 phrase pairs from the pool of non-misaligned phrases, observing an average JPO adequacy score of 4.82, implying the amateur and crowd-sourced translations are high quality.

\subsubsection{Machine translation performance }
We also evaluated JESC with downstream Machine Translation performance, using the TensorFlow and Sequence-to-Sequence frameworks \cite{abadi2016tensorflow,britz2017massive,lison2016opensubtitles2016}. 
We used a 4-layer bidirectional LSTM encoder and decoder with 512 units, as well as dot-product attention \cite{luong2015effective}. We applied Dropout at a rate of 0.2 to the input of each cell, and optimized using Adam and a learning rate of 0.0001 \cite{kingma2014adam}. We used a batch size of 128, and train for 10 epochs. For each experiment, we preprocess the data using learned subword units\footnote{\texttt{https://github.com/google/sentencepiece}} \cite{sennrich2015neural} for a shared vocabulary of 16,000 tokens.

In addition to evaluating JESC, we trained and tested on the ASPEC corpus of \cite{nakazawa2016aspec} which consists of scientific abstracts (3M examples), the Kyoto Wiki Corpus (KWC) of \cite{chu2014constructing} which consists of translated Wikipedia articles (0.5M examples), and the OpenSubs corpus of \cite{lison2016opensubtitles2016} which is the closest analog to JESC and consists of 1M professionally-made and automatically aligned captions.

\begin{table}[H]
\centering
\begin{tabular}{l|llll}
Train/Test & ASPEC & KWC & OpenSubs & JESC \\ \hline
ASPEC      & {\bf 36.23} & 15.42 & 3.45     & 3.81 \\
KWC      & 5.30  & {\bf 8.61}  & 2.31     & 2.22 \\
OpenSubs   & 0.2   & 0.7   & {\bf 10.01}     & 6.3  \\
JESC       & 2.35  & 3.71  & 8.8      & {\bf 14.21}
\end{tabular}
\caption{Machine translation results (BLEU).}
\label{table:mt-results}
\end{table}

Even though KWC consists of high quality and human-made translations, we find that it underperforms due to the small size of the dataset (Table \ref{table:mt-results}). Similarly, we find that JESC's large size helps it outperform OpenSubs in both in-domain BLEU and out-of-domain generalization.

\section{Conclusion}

We introduced JESC, a large-scale parallel corpus for the Japanese-English language pair. JESC is (1) the largest publicly available Japanese-English corpus to date, (2) a corpus that covers the underrepresented domain of conversational speech, and (3) to the extent of these authors knowledge, the only large-scale parallel corpus to support multi-reference BLEU evaluation. Our experimental results suggest that these data are a high quality and novel challenge for today's machine translation systems. By releasing these data to the public, we hope to increase the colloquial abilities of today's MT systems, especially for the Japanese-English language pair.

\section{Bibliographical References}
\label{main:ref}

\bibliographystyle{lrec}
\bibliography{xample}

\end{document}